\pdfoutput=1
\documentclass[10pt,twocolumn,letterpaper]{article}

\usepackage{cvpr}              

\usepackage{bbding}
\usepackage{graphicx}
\usepackage{booktabs}
\usepackage{threeparttable}
\usepackage[table]{xcolor}
\usepackage{tabularx}
\usepackage{hyperref}
\usepackage{xcolor}
\usepackage{multirow}

\usepackage[linesnumbered,ruled,vlined,nofillcomment]{algorithm2e}
\SetKwInput{KwInit}{Init}
\SetKwInput{KwInput}{Input}
\SetKwProg{Fn}{Function}{:}{}
\newcommand{\dataset}{SOK-Bench\xspace}

\definecolor{cvprblue}{rgb}{0.21,0.49,0.74}
\usepackage[normalem]{ulem}
\usepackage{hyperref}

\usepackage{xcolor,pifont}
\newcommand*\colourcheck[1]{%
  \expandafter\newcommand\csname #1check\endcsname{\textcolor{#1}{\ding{52}}}%
}
\colourcheck{blue}
\colourcheck{green}
\colourcheck{red}

\def\paperID{3757} 
\def\confName{CVPR}
\def\confYear{2024}

\title{\dataset : A Situated Video Reasoning \\ Benchmark with Aligned Open-World Knowledge}

\author{
\textbf{Andong Wang}\thanks{The authors contributed equally to the work.} \textsuperscript{1}, \textbf{Bo Wu}\footnotemark[1] \textsuperscript{2}, \textbf{Sunli Chen}\textsuperscript{3}, \textbf{Zhenfang Chen}\textsuperscript{2}, \textbf{Haotian Guan}\textsuperscript{1} \\ \textbf{Wei-Ning Lee}\textsuperscript{1}, \textbf{Li Erran Li}\textsuperscript{4}, \textbf{Chuang Gan} \textsuperscript{5,2} \\
\textsuperscript{1}The University of Hong Kong, \textsuperscript{2}MIT-IBM Watson AI Lab, \\
\textsuperscript{3}Tsinghua University, \textsuperscript{4}AWS AI, \textsuperscript{5}UMass Amherst\\
}

\begin{document}

\newcolumntype{C}[1]{>{\centering\arraybackslash}p{#1}}

\maketitle
\begin{abstract}
Learning commonsense reasoning from visual contexts and scenes in real-world is a crucial step toward advanced artificial intelligence. However, existing video reasoning benchmarks are still inadequate since they were mainly designed for factual or situated reasoning and rarely involve broader knowledge in the real world.
Our work aims to delve deeper into reasoning evaluations, specifically within dynamic, open-world, and structured context knowledge. 
We propose a new benchmark (\dataset), consisting of 44K questions and 10K situations with instance-level annotations depicted in the videos. The reasoning process is required to understand and apply situated knowledge and general knowledge for problem-solving.
To create such a dataset, we propose an automatic and scalable generation method to generate question-answer pairs, knowledge graphs, and rationales by instructing the combinations of LLMs and MLLMs. 
Concretely, we first extract observable situated entities, relations, and processes from videos for situated knowledge and then extend to open-world knowledge beyond the visible content. 
The task generation is facilitated through multiple dialogues as iterations and subsequently corrected and refined by our designed self-promptings and demonstrations.
With a corpus of both explicit situated facts and implicit commonsense, we generate associated question-answer pairs and reasoning processes, finally followed by manual reviews for quality assurance.
We evaluated recent mainstream large vision-language models on the benchmark and found several insightful conclusions.
For more information, please refer to our benchmark at \url{www.bobbywu.com/SOKBench}. 

\end{abstract}

\section{Introduction}
\label{sec:intro}
\begin{figure*}[th]
  \setlength{\abovecaptionskip}{1pt}
  \centering
  \vspace{-0.5em}
  \includegraphics[width=18cm]{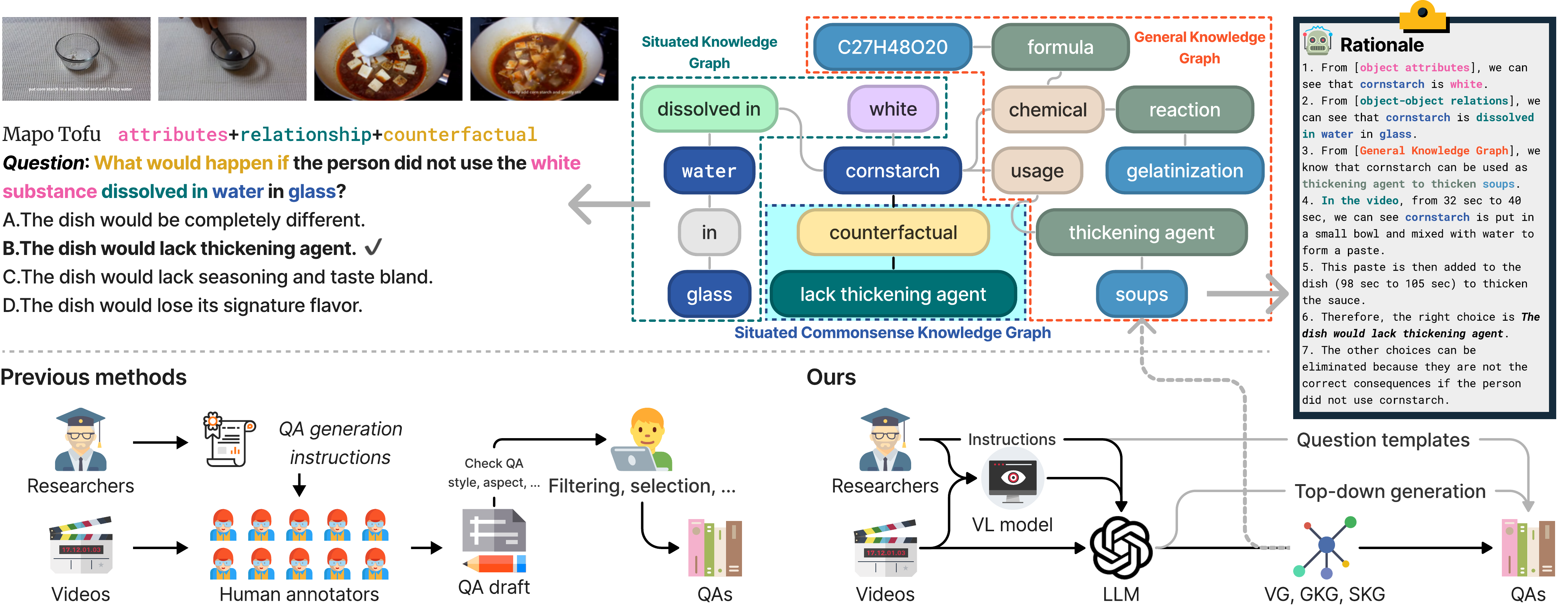}
  \caption{Overview. Instead of using crowd-sourced methods, we design a synthesis pipeline to create the benchmark by leveraging use LLMs and VLLMs, improving efficiency and ensuring consistency. The method helps to automatically generate high-quality question-answers (QAs) and focusing the desirable purposes for evaluating the model's ability. To generate data aligned with open-world knowledge, we propose to connect situation, general knowledge, and situated commonsense and produced three types of associated knowledge graphs (refer to the subsection~\ref{sec:method_situated_knowledge_graph},~\ref{sec:method_general_knowledge_graph}, and~\ref{sec:method_situated_commonsense_knowledge_graph}). Specifically, it makes more precise inferences based on situational facts and essential commonsense knowledge by aligning with the bottom-up or top-down goals, the reasoning process from Q to A is able to demonstrate explicitly.}
  \vspace{-1em}
\label{fig:teaser}
\end{figure*}

The evolutions of Large Language Models (LLMs)~\cite{radford2019language,ouyang2022training,touvron2023llama,openai2023gpt} and Multimodal Large Language Models (MLLMs)~\cite{li2023blip,zhu2023minigpt,liu2023visual,li2023videochat,zhang2023video,su2023pandagpt,luo2023valley} mark a significant milestone in artificial intelligence. 
These models are making remarkable strides across various domains, exhibiting better capabilities in perceptual, generative, and comprehensive tasks~\cite{zheng2023judging,bubeck2023sparks}. 
As models like ChatGPT~\cite{ouyang2022training,openai2023gpt} and successors continue to improve in scale, the possibility to capture commonsense knowledge has also seen obvious advancements. 
Hence, building models with real-world knowledge is becoming more promising now which was a long-standing but difficult challenge before~\cite{tandon2018,chadha2021ireason}. 

Besides novel models, previous efforts have been directed towards evaluating multimodal commonsense reasoning, transitioning from purely language-based models to those incorporating visual comprehension, such as Visual Question Answering (VQA)~\cite{antol2015vqa,jang2017tgif,lei2018tvqa,mun2017marioqa} or Visual Commonsense Reasoning (VCR)~\cite{johnson2017inferring,johnson2017clevr,hudson2019gqa,yi2019clevrer,wu2021star,grunde2021agqa}. 
These tasks have an underlying assumption that the given visual and language inputs (e.g., images, videos, questions, etc.) contain most of the required information. 
However, relying on the given task context does not fully reflect real-world complexities, narrowing the scope of their reasoning evaluation.
More recently, the focus has shifted towards developing tasks that require external knowledge for commonsense reasoning~\cite{zellers2019recognition,marino2019ok,schwenk2022okvqa} and consider the integration of factual~\cite{wang2017fvqa}, evidential~\cite{li2022from}, or external knowledge sources~\cite{zellers2019recognition,marino2019ok,schwenk2022okvqa,xiao2021next,gupta2022newskvqa} into evaluations. 
They primarily provide visual information and external information on static images for reasoning. Thus, this way overlooks several crucial natures such as spatio-temporal, causal, or dynamic processes of real-world activities and events, an aspect crucial for truly understanding and interacting with our environment. 
Also, the knowledge is written and concluded by free-form descriptions from crowd-source ways~\cite{zellers2019recognition,marino2019ok,gupta2022newskvqa,xiao2021next,schwenk2022okvqa}, presenting its own set of limitations. Consistency in descriptions provided by annotators is difficult to maintain~\cite{marino2019ok,zellers2019recognition}, and there are multiple risks of the method being influenced by distinct descriptive or empirical biases~\cite{johnson2017clevr,hudson2019gqa,wu2021star,grunde2021agqa}. 


Our research aims to delve deeper into the realm of commonsense reasoning, specifically within dynamic, open-world, and structured contexts. 
We build Situated Open-World Commonsense Reasoning (\dataset), a novel benchmark consists of over 44K questions with answers and situated commonsense knowledge graphs, covers over 12 types of questions, and sources from about 10K dynamic situations in real-world activities. 
The models are required to produce appropriate inferences by leveraging both facts within situations and the necessary commonsense or background knowledge. 

Thus we propose the Situated Commonsense Graph to connect and integrate the required knowledge of the reasoning process where situated knowledge and general knowledge are structured, compositional, and aligned. Compared with other datasets in Figure~\ref{tab:dataset_comparison}, our reasoning benchmark has diverse characteristics including instance annotations, compositional generation, and structural alignment of the situated open-world knowledge and rationales.
Moreover, we propose a scalable and automatic pipeline for both task generation and annotation. The typical way to generate evaluation task instructions is crowd-source human annotation~\cite{antol2015vqa,zellers2019recognition} or adversarial filtering~\cite{zellers2018swag}. However, collecting and locating high-quality situated commonsense data are not trivial since considering situation, question, answer and their inner relations requires experienced annotators, huge cost, and high consistency. 
Instead, we generate evaluation tasks, questions with answers, and reasoning processes, which integrate situated knowledge, general knowledge, and underlying reasoning commonsense goals or logics.
We design appropriate instructions and compositions for prompts and define question types, goals to enhance generation quality, logic, and rationality, as shown in Figure~\ref{fig:dataset_stat} (a).
The iterative generation process automatically utilizes the interactions of LLMs or MLLMs in multi-turns, as shown in Figure~\ref{fig:graph_gen_pipeline}.
In experiments, we evaluate the mainstream LLMs and VideoLLMs as representatives for the proposed reasoning benchmark. 
Although the models have trained on web-scale data or adopted pre-trained foundation models as basis, the experiments indicate significant room for future improvement.
We also provide comprehensive ablation comparisons and analysis for generation processes.



\begin{itemize}
    \item We create \dataset, a novel benchmark to evaluate situated and open-world commonsense reasoning in videos with compositionality, temporality, and causality.
    \item We propose a scalable method to generate video question-answering pairs for reasoning through iterative conversations with LLMs and MLLMs. We designed diverse prompt compositions and a self-prompting strategy to construct the video knowledge, commonsense knowledge, situated commonsense knowledge graphs, \etc.
    \item We evaluated representative LLMs and MLLMs on the proposed benchmark in different settings and found that existing LLMs and MLLMs still perform inferior on situated open-knowledge reasoning in videos, which further proves our new dataset's research value.
\end{itemize}

\begin{table*}[t]
    \setlength{\abovecaptionskip}{0pt}
    \setlength{\belowcaptionskip}{0pt}
    \setlength{\tabcolsep}{3pt} 
    \caption{Comparison between the proposed benchmark with existing datasets. }
    \scriptsize
    \centering
    \begin{threeparttable}
    \resizebox{1\linewidth}{!}{
    \begin{tabular}{l|cccc|c|cc|c|cc}
    \toprule
     & \multicolumn{3}{c}{Generation} & QA & Knowledge & \multicolumn{2}{c}{Situated Knowledge} & General & \multicolumn{2}{c}{Situated Commonsense Knowledge} \\ 
     & Approach & Real-world & Scalability & Rationale & Representation & Grounding & Dynamic & Knowledge & Existence & Compositionality \\ 
    \cmidrule(lr){1-11}
    CLEVR~\cite{johnson2017clevr} & Rule & \textcolor{red}{\XSolidBrush} & \greencheck & \greencheck & program & \greencheck & \greencheck & single-source & \greencheck & \greencheck \\
    CLEVRER~\cite{yi2019clevrer} & Rule & \textcolor{red}{\XSolidBrush} & \greencheck & \greencheck & program & \greencheck & \greencheck & single-source & \greencheck & \greencheck \\
    EgoTV~\cite{hazra2023egotv} & Rule & \textcolor{red}{\XSolidBrush} & \greencheck & \greencheck & program & \greencheck & \greencheck & single-source & \textcolor{red}{\XSolidBrush} & \textcolor{red}{\XSolidBrush} \\
    STAR~\cite{wu2021star} & Rule & \greencheck & \greencheck & \greencheck & graph & \greencheck & \greencheck & single-source & \textcolor{red}{\XSolidBrush} & \textcolor{red}{\XSolidBrush} \\
    AGQA~\cite{grunde2021agqa} & Rule & \greencheck & \greencheck & \greencheck & graph & \greencheck & \greencheck & single-source & \textcolor{red}{\XSolidBrush} & \textcolor{red}{\XSolidBrush} \\
    VCR~\cite{zellers2019vcr} & Manual & \greencheck & \textcolor{red}{\XSolidBrush} & \greencheck & description & \greencheck & \greencheck & multi-source & \greencheck & \textcolor{red}{\XSolidBrush} \\
    VIOLIN~\cite{liu2020violin} & Manual & \greencheck & \textcolor{red}{\XSolidBrush} & \textcolor{red}{\XSolidBrush} & description & \greencheck & \greencheck & single-source & \greencheck & \textcolor{red}{\XSolidBrush} \\
    NExT-QA~\cite{xiao2021next} & Manual & \greencheck & \textcolor{red}{\XSolidBrush} & \textcolor{red}{\XSolidBrush} & description & \textcolor{red}{\XSolidBrush} & \greencheck & single-source & \greencheck & \textcolor{red}{\XSolidBrush} \\
    Causal-VidQA~\cite{li2022from} & Manual & \greencheck & \textcolor{red}{\XSolidBrush} & \textcolor{red}{\XSolidBrush} & description & \textcolor{red}{\XSolidBrush} & \greencheck & single-source & \greencheck & \textcolor{red}{\XSolidBrush} \\
    V2C~\cite{fang-etal-2020-video2commonsense} & Manual & \greencheck & \textcolor{red}{\XSolidBrush} & \greencheck & graph & \textcolor{red}{\XSolidBrush} & \greencheck & multi-source & \greencheck & \textcolor{red}{\XSolidBrush} \\
    OK-VQA~\cite{marino2019ok} & Manual & \greencheck & \textcolor{red}{\XSolidBrush} & \textcolor{red}{\XSolidBrush} & description & \textcolor{red}{\XSolidBrush} & \textcolor{red}{\XSolidBrush} & multi-source & \greencheck & \textcolor{red}{\XSolidBrush} \\
    A-OKVQA~\cite{schwenk2022okvqa} & Manual & \greencheck & \textcolor{red}{\XSolidBrush} & \greencheck & description & \textcolor{red}{\XSolidBrush} & \textcolor{red}{\XSolidBrush} & open-world & \greencheck & \textcolor{red}{\XSolidBrush} \\
    \rowcolor{gray!20} Ours & Auto, Rule & \greencheck & \greencheck & \greencheck & graph & \greencheck & \greencheck & open-world & \greencheck & \greencheck \\
    \bottomrule
    \end{tabular}
    }
    \end{threeparttable}
    \label{tab:dataset_comparison}
\end{table*}



\section{Related Works}
\label{sec:related_works}
\subsection{Video Question Answering}
\noindent{\textbf{Video Question Answering.}}
Our work is closely related to video question answering, which requires a model to watch the video and answer questions related to the video's content~\cite{lei2018tvqa,wu2021star,xiao2021next,li2022from,grunde2021agqa,yu2023self,chen2022comphy}. Most of these existing benchmarks ask questions about visual attributes, human actions, and activities, or physical intuition~\cite{yi2019clevrer,chen2022comphy, patel2022cripp}. However, none of them has focused on studying events in videos with situated knowledge, open-domain knowledge, and explicit multi-step reasoning rationales.

\noindent{\textbf{Commonsense Question-Answering.}}
Our research is aligned with the field of commonsense question answering. It requires models to make use of commonsense to answer questions. It was first studied in the field of natural question answering~\cite{yang2018hotpotqa,nguyen2016ms} without any visual context as input. Later, people evaluated commonsense understanding in the context of visual question answering~\cite{marino2019ok,schwenk2022okvqa,park2020visualcomet,chang2022webqa}. Given a static image, such benchmarks require models to answer questions based on the visual context of the image and its associated commonsense knowledge. There are some benchmarks studying domain-specific commonsense like physics~\cite{ates2020craft} and social events~\cite{zadeh2019social}. However, we are the first benchmark to study reasoning with situated and open-world knowledge in dynamic situations with dense annotations like knowledge graphs and rationales. Table~\ref{tab:dataset_comparison} summarizes the differences compared with previous benchmarks.

\noindent{\textbf{LLM for Data Annotation.}}
Our work is situated within the domain of using large language models for data annotation~\cite{ding2022gpt,pei2023gpt}. Ding \etal~\cite{ding2022gpt} examined the effectiveness of GPT-3 as a data annotator for NLP tasks by comparing it with traditional annotation methods and analyzing its performance across various tasks.  Pei~\etal~\cite{pei2023gpt} introduced a self-supervised GPT annotation method using a generating-recovering paradigm, leveraging one-shot learning for efficient data-to-summary annotation, and evaluated its performance using alignment scores and human feedback reward networks. Both of these two works only focus on the natural language side. In this work, we propose to prompt LLMs to cooperate with different vision models~\cite{li2023blip,zhu2023minigpt} to annotate videos and generate question-answer pairs related to video content and open knowledge.

\noindent{\textbf{Vision-Language Models.}}
Recently, there has been great interest in large pre-trained vision-language models~\cite{zhang2023video,su2023pandagpt,li2023videochat,luo2023valley,li2023blip,zhu2023minigpt} (VLMs). Early models~\cite{li2023blip,zhu2023minigpt} only evaluate their performance on inputs with static images and natural language. There are also some models~\cite{luo2023valley} evaluating performance their performance on videos. However, these VLMs either only showed some qualitative examples~\cite{zhang2023video,su2023pandagpt,li2023videochat} or evaluates on traditional video question answering benchmarks~\cite{xu2017video,xu2016msr,yu2019activitynet}. However, ~\cite{xu2017video,xu2016msr,yu2019activitynet} are not ideal benchmarks for commonsense reasoning since they mainly focus on the perception side and require no external knowledge to answer the questions. It remains an open question as to how to evaluate VLMs' performance in understanding situations in videos and reasoning about open-domain knowledge with commonsense. 

\section{Situated Commonsense Reasoning}
\label{sec:benchmark}

\begin{figure*}[h]
  \setlength{\abovecaptionskip}{1pt}
  \centering
  \includegraphics[width=18cm]{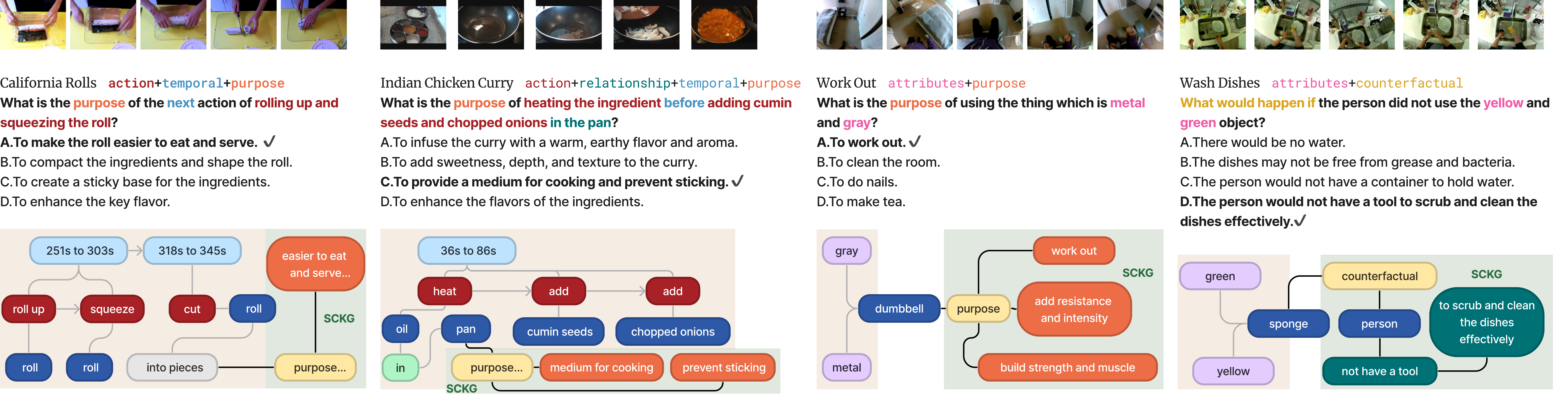}
  \caption{SOK-Bench data examples. Each QA pair corresponds to a video clip (\eg, a video clip showing how to cook \emph{California Rolls}) and a type of situated commonsense knowledge (\eg, \emph{action + temporal + purpose}). For each question, we provide four options, the correct choice, and the associated situated commonsense graphs.}
\label{fig:qa_examples}
\end{figure*}

\subsection{Benchmark Overview}
\begin{figure}[t]
  \setlength{\abovecaptionskip}{1pt}
  \centering
  \includegraphics[width=8.5cm]{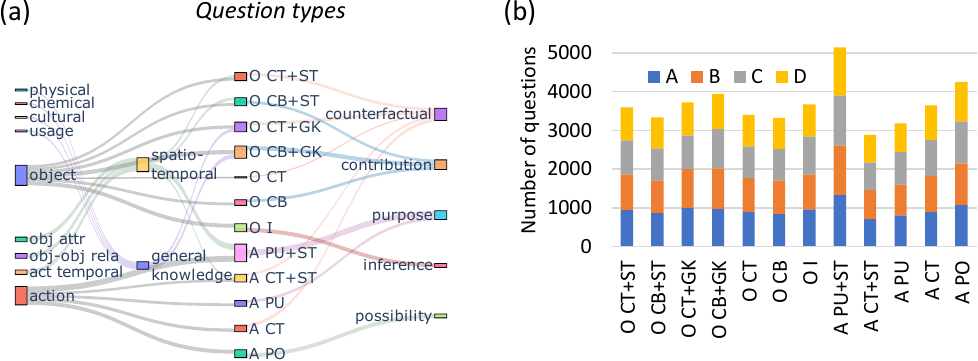}
  \caption{\textbf{(a)} Sankey diagram of the 12 question types. \textbf{(b)} Answer distribution among options for each question type. Meaning of abbreviations: \textbf{O}: Object; \textbf{A}: Action; \textbf{CT}: Counterfactual; \textbf{CB}: Contribution; \textbf{PU}: Purpose; \textbf{I}: Inference; \textbf{PO}: Possibility; \textbf{ST}: Spatiotemporal; \textbf{GK}: General knowledge. Notably, the ``Spatiotemporal'' includes ``obj attributes'', ``obj-obj relations'', ``obj attribute + obj-obj relation'', and ``before/after action'' (see Section~\ref{sec:method_qa_gen}).}
\label{fig:dataset_stat}
\end{figure}

\noindent{\textbf{Statistics.}}
Our benchmark consists of 44K questions and answers for reasoning with situated open-world knowledge, accompanied by multiple-choice options, and 10K video clips. Each question-answer (QA) pair links to a hypergraph which is generated from the Situated Knowledge Graph, General Knowledge Graph, and Situated Commonsense Knowledge Graph (refer to the below sections and the algorithms in the Supplementary). The aligned graphs effectively showcase the relations between situated knowledge and general knowledge. Figure~\ref{fig:qa_examples} presents four QA examples involving different types situated commonsense knowledge. We also present a detailed rationale that elucidates the sequence of reasoning steps that bridge the gap from each question to its corresponding answer, offering a clearer understanding of the underlying thought process (see Figure~\ref{fig:teaser}).
Figure~\ref{fig:dataset_stat} showed the 12 question types.
Each question is accompanied by a direct answer and a set of four multiple-choice options. The dual format ensures versatility in response assessment, as depicted in Figure~\ref{fig:dataset_stat}.


\noindent{\textbf{Benchmark Generation.}}
We present a new method for creating our benchmark automatically in a structured, controllable, and scalable way. Our approach simplifies manual and cumbersome processes and involves four streamlined stages: 1) Extracting observable content from the videos; 2) Compiling relevant commonsense knowledge exhaustively; 3) Aligning the content of the situations with this commonsense knowledge to reveal underlying logical connections and implications; 4) Formulating questions and answers by integrating the gathered information.

We utilize in-context learning, few-shot demonstrations, and multiple rounds of interactions to prompt LLMs and MLLMs for the generation. They are guided by a set of prompt templates to produce the proposed knowledge graphs.
For the final stage, these graphs aid in the creation of question templates, either through manual construction or by directing the LLM to generate question-answer pairs (QAs). This strategy ensures a direct correlation between the QAs, the graphs, and the rationale that maps the reasoning from questions to answers.
We've taken measures to ensure an unbiased distribution of answer options and have employed both LLM and human annotators to identify and correct any grammatical errors. 


\begin{figure*}[t]
  \setlength{\abovecaptionskip}{1pt}
  \centering
  \includegraphics[width=18cm]{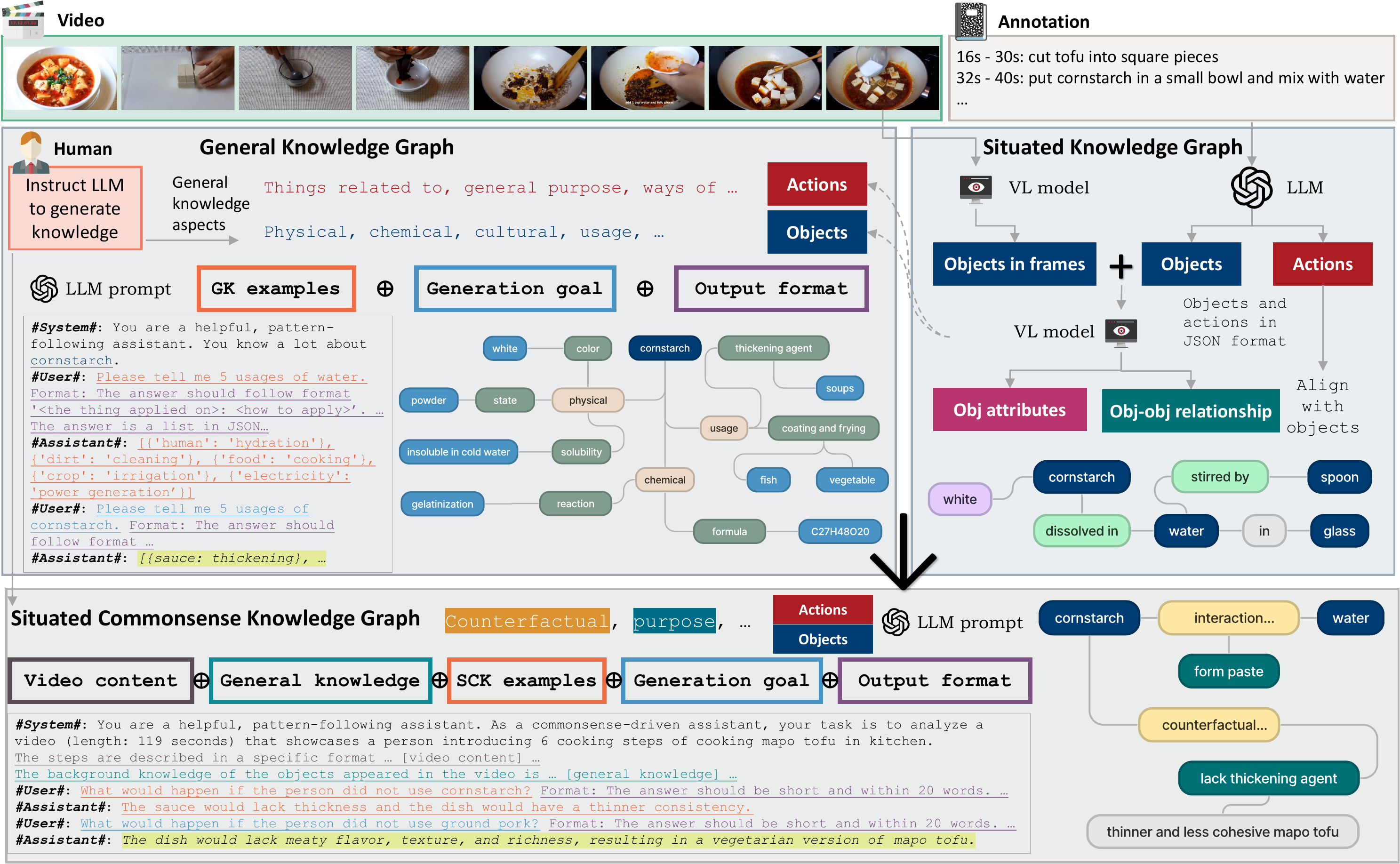}
  \caption{Generation pipeline of Situated Knowledge Graph (SKG, see Section~\ref{sec:method_situated_knowledge_graph}), General Knowledge Graph (GKG, see Section~\ref{sec:method_general_knowledge_graph}), and Situated Commonsense Knowledge Graph (SCKG, see Section~\ref{sec:method_situated_commonsense_knowledge_graph}).}
\label{fig:graph_gen_pipeline}
\end{figure*}

\subsection{Situated Knowledge Graph}
\label{sec:method_situated_knowledge_graph}






We derive dynamic situations from videos, which provide insights into event temporality, causality, and dynamics~\cite{buch2022}. 
Recent research has increasingly represented objects and relationships in real-world dynamic situations~\cite{wu2021star,grunde2021agqa,urooj2023learning} and visual scenes~\cite{ji2020action,zellers2019recognition,hudson2019gqa} using graphs.
Building on this, we expand upon traditional visual graphs to introduce our Situated Knowledge Graph. This includes more comprehensive information derived from dynamic situations, such as people and objects, their attributes, relationships, and actions. 

Our process begins by generating object and action entities, which are identified based on descriptions provided with source videos (\eg, shown in Figure~\ref{fig:graph_gen_pipeline}). 
These descriptions segment dynamic situations into distinct activity moments, associating objects $\{o_{i}\}$ and actions $\{a_{t}\}$ with specific timestamps $\{y_{t}\}$.
Given original descriptions are human-written and lack a standardized format, we employ Large Language Models (LLMs) to parse the descriptive text. The LLMs systematically convert the sentences into organized lists of objects and actions, along with their corresponding timestamps. This approach streamlines the extraction of key elements from the narrative text.
For some objects that are observable in video frames but were not annotated, we use BLIP2~\cite{li2023blip} to recognize objects in frames and add them to the initial object list $\{o_{i}\}$.
Next, we extract object attributes and spatial relationships between objects from selected keyframes within the videos. To achieve this, we utilize MiniGPT4~\cite{openai2023gpt} to articulate these object attributes and their spatial interrelations. 
Using a designated \textit{goal} prompt, our method efficiently generates object attributes for visual, physical, and chemical properties and relationships (\eg, \#Prompt\#: Please describe the tofu in the image in terms of size, shape, color, and texture in one short sentence within 10 words. \#Model\#: The tofus are small, white, sliced squares.). 
Finally, we align the timestamps of frames with those of actions. When an action's timestamp coincides with the time at which an object appears in a frame, we link this action and object by adding an \textit{(action, object)} edge to the situated knowledge graph.
Additionally, for each action,
we build an additional edge \textit{(timestamp, action)} to indicate the action sequence.

\subsection{General Knowledge Graph}
\label{sec:method_general_knowledge_graph}

We construct the general knowledge graph beginning from the object $o_{i}$ or action $a_{t}$ as initial nodes. Taking ``tofu'' as a node example (a food product prepared by coagulating soy milk and then pressing the resulting curds into solid white soft toufu blocks), prompting an LLM with a goal query (\eg, ``Tell me some common knowledge about tofu'') is a basic approach. While this will generate descriptions with relevant knowledge, the resulting content lacks specificity and completeness, and always misses key points (e.g., the color of tofu). 
Thus, we propose multiple aspects for knowledge generation (\eg, \textit{usage}, \textit{physical}, and \textit{culture}, etc) and collaborate with ``goal'' to make it with concrete descriptions.
The direct challenges of LLM outputs are verbose, unstructured, or repetitive (\eg, `` Tofu is a popular food ...'' or ``Tofu is a versatile ingredient that ...''). To address this, we use \textit{general knowledge examples} with the predefined \textit{output format} to extend the knowledge graph edges or nodes.
Moreover, relying solely on the format prompt can lead to format inconsistencies in outputs, such as missing quotation marks (\eg, \{``tofu'': ...\{``physical'': ``color'': ``white\}\} that cause parsing failures. 
Thus, we utilize a prompt triplet $\langle$\textit{general knowledge examples} $\pi$, \textit{generation goal} $\omega$, \textit{output format} for GKGs (Figure~\ref{fig:graph_gen_pipeline}). By employing both \textit{general knowledge examples} and \textit{output format}, we ensure the generated knowledge is not only of high quality but also consistently parseable.
Here are the generated outputs for cornstarch usages: [\{``sauce'': ``thickening''\}, \{``batter'': ``making crispy''\}, \{``body powder'': ``absorbing moisture''\}, \{``starch'': ``making''\}, \{``playdough'': ``creating homemade''\}]. 

\subsection{Situated Commonsense Knowledge Graph}
\label{sec:method_situated_commonsense_knowledge_graph}

We propose Situated Commonsense Knowledge Graph (SCKG) by integrating information from the previously established SKG and GKG for specific dynamic situations. 
To illustrate the significance and complexity of generating situated commonsense knowledge, consider a straightforward example: thinking about the impact of omitting cornstarch in the preparation of mapo tofu. Merely knowing that cornstarch is a powdery substance (from general knowledge) is insufficient. We need to combine this with its situated use from the SKG (\eg, ``The person mixes cornstarch with water and adds the mixture to the soup.'') and its general properties from the GKG (\eg, ``Cornstarch can be used as a thickening agent.''). The result is ``The sauce would be less thick, and the dish would have a thinner consistency.''

We initially thought that a four-element prompt, $\langle$\textit{video content} $g_{v}$, \textit{general knowledge} $k_{g}$, \textit{generation goal} $\omega$, \textit{output format} $\phi$$\rangle$, would work for generating situated commonsense knowledge. Our test indicates that the model, while generating correct responses, often provides answers that are too general. For instance, in response to a question about the impact of not using ground pork in mapo tofu, the model might simply reply, ``The dish would lack flavor and texture'' which, although accurate, lacks detail.

Directly providing examples to instruct the LLM to achieve a specific level of detail isn't practical due to the dependency of knowledge on the varying video content. To address this issue, we propose a method called Few-Shot Self-Prompting.


\textbf{Few-Shot Self-Prompting} uses previously generated situated commonsense knowledge as examples. The prompt consists of five elements: $\langle$\textit{video content} $g_{v}$, \textit{general knowledge} $k_{g}$, \textit{situated commonsense knowledge examples} $\pi$, \textit{generation goal} $\omega$$\rangle$. The details of the algorithm can be found in the Supplementary Materials.

Figure~\ref{fig:graph_gen_pipeline} demonstrates that the situated knowledge examples are constructed using knowledge about the consequences of not using a specific ingredient (\eg, not using \emph{cornstarch} during cooking). Over time, this process leads to more concrete and specific situated commonsense knowledge. The final result, for example, would be, ``The dish would lack meaty flavor, texture, and richness, resulting in a vegetarian version of mapo tofu.''

Our Few-Shot Self-Prompting technique enables us to explore various situated commonsense knowledge perspectives, such as considering counterfactual scenarios, understanding the purpose of an action, and recognizing an object's contribution. Our experiments show that just two iterations of knowledge generation can produce high-quality results, ultimately leading to the creation of the SCKG.

\subsection{Question and Answer Generation}
\label{sec:method_qa_gen}

Using the three graphs, we create question-answer pairs to test the model's ability to make accurate inferences based on situational facts and essential commonsense knowledge. We can use two approaches for this: \textbf{(1)} Manually create question-answer templates in a bottom-up manner, designing question templates and providing answers based on the graphs. \textbf{(2)} Automatically generate questions using a LLM in a top-down manner.

\subsubsection{Bottom-up QA Generation}
\label{sec:method_rule_based_qa_gen}

We manually design question templates in multi hog way combing both ``situation'' and ``commonsense'' (please see the details of template types in Supplementary Materials). For one object / action, we find the connected edges in the three graphs. Next, we design question templates that jump from edges in Situated Knowledge Graph or General Knowledge Graph to those in Situated Commonsense Knowledge Graph. Concretely, for example, the template can be ``What would happen if the person did not use the \textit{\textless{}obj attribute\textgreater{}} and \textit{\textless{}obj-obj relation\textgreater{}}?''(see one concrete question example concerning ``cornstarch'' in Figure~\ref{fig:teaser}). The model needs to first do spatiotemporal reasoning to identify the object by \textit{\textless{}obj attribute\textgreater{}} and \textit{\textless{}obj-obj relation\textgreater{}} (\eg, ``white substance'' and ``dissolved in water in glass''). Next, the model needs to do situated commonsense reasoning to understand the consequence if that object was not used. In such way, we can comprehensively assess model's situated commonsense reasoning ability. Note that we also present some easier templates (\eg, ``vanilla obj counterfactual'') which are single hog.

Apart from the question and the correct answer, consistent with VCR~\cite{zellers2019vcr}, we generate wrong answers that are relevant but sufficiently dissimilar from the correct answers to minimize shortcut risk (shown in Figure~\ref{fig:qa_examples}). Incorrect options are chosen from the same category of contextual knowledge related to other objects or actions. Also, we ensure an even distribution of the four choices (Figure~\ref{fig:dataset_stat}(b)).

\subsubsection{Top-down QA Generation}
\label{sec:method_top_down_qa_gen}

Bottom-up QA generation allows for control over question quality and difficulty but may not create diverse question types. To address this, we suggest a top-down approach for QA generation. We design a structured prompt with five elements: $\langle$\textit{video content}, \textit{integrated graph}, \textit{QA examples}, \textit{generation goal}, \textit{output format}$\rangle$. The integrated graph combines SKG, GKG, and SCKG. The generation goal directs the LLM to create multi-hog questions based on multiple edges from the integrated graph. This prompt design keeps a strong connection between generated QAs and graphs, reducing the chance of model hallucination compared to basing QAs only on video content.

However, the top-down approach has drawbacks, and it's resource-intensive and slow. With the three graphs involved, it takes a few seconds to generate one question using the top-down method, while the bottom-up method generates the entire benchmark in the same time. In this paper, we mainly use the bottom-up method to create the benchmark, but we discuss top-down generation case studies in the Supplementary.

\subsection{Quality Valiation}
We invite human helpers to assess the quality by inspecting a subset of the graphs and QAs. We find 93.08\% of QA pairs are valid. Please see the Supplementary for details.



\subsection{Data Source}
We created the benchmark based on videos of the two public video datasets about daily activities.


\noindent\textbf{YouCook2}~\cite{zhou2018towards} is a large set of instructional cooking videos from YouTube. It includes over 2,000 videos across 89 recipe categories, with each video annotated with detailed step-by-step instructions.

\noindent\textbf{HOMAGE}~\cite{rai2021home}  contains 1,752 video clips featuring 75 daily human activities in home environments. The dataset is thoroughly annotated with action labels, object information, and spatial-temporal relationships



\section{Experiments}
\label{sec:experiments}

\begin{table*}[!htb]
    \setlength{\abovecaptionskip}{1pt}
    \caption{Accuracy of baseline models in multiple-choice setting. The question types are represented by numbers from 1 to 12:
1. Object Counterfactual + Spatiotemporal (OCT+ST);
2. Object Contribution + Spatiotemporal (OCB+ST);
3. Object Counterfactual + General Knowledge (OCT+GK);
4. Object Contribution + General Knowledge (OCB+GK);
5. Object Counterfactual (OCT);
6. Object Contribution (OCB);
7. Object Inference (OI);
8. Action Purpose + Spatiotemporal (APU+ST);
9. Action Counterfactual + Spatiotemporal (ACT+ST);
10. Action Purpose (APU);
11. Action Counterfactual (ACT);
12. Action Possibility (APO).}
    \small
    \centering
    \begin{threeparttable}
    \resizebox{1\linewidth}{!}{
    \begin{tabular}{llllllllllllll}
    \toprule
    \multirow{2}{*}{Models} & \multicolumn{7}{c}{Questions concerning objects}        & \multicolumn{5}{c}{Questions concerning   actions} & \multirow{2}{*}{Overall} \\
    \cmidrule(lr){2-8}\cmidrule(lr){9-13}
                            & 1 & 2 & 3 & 4 & 5 & 6 & 7 & 8 & 9 & 10 & 11 & 12  &                          \\
    \cmidrule(lr){1-14}
    blind ChatGPT~\cite{OpenAI2023ChatGPT} & 0.224& 0.318& 0.441& 0.275& 0.651& 0.488& 0.206& 0.136& 0.135& 0.574& 0.729& 0.189& 0.365 \\
    GPT4v~\cite{openai2023gpt4}          & 0.000 & 0.600 & 0.600 & 0.333 & 0.800 & 1.000 & 0.400 & 0.300 & 0.222 & 0.778 & 0.885 & 0.600 & 0.539 \\
    Video-LLaMa~\cite{zhang2023video} & 0.251 & 0.265 & 0.254 & 0.258 & 0.301 & 0.273 & 0.319 & 0.267 & 0.257 & 0.255 & 0.257 & 0.242 & 0.264 \\
    PandaGPT~\cite{su2023pandagpt} & 0.264 & 0.281 & 0.286 & 0.281 & 0.373 & 0.335 & 0.368 & 0.236 & 0.223 & 0.344 & 0.445 & 0.350 & 0.312 \\
    Ask Anything~\cite{li2023videochat} & 0.256 & 0.362 & 0.441 & 0.404 & 0.600 & 0.543 & 0.276 & 0.288 & 0.180 & 0.713 & 0.729 & 0.65 & 0.455\\
    Video-ChatGPT~\cite{li2023videochat} & 0.309 & 0.206 & 0.274 & 0.210 & 0.394 & 0.311 & 0.274 & 0.277 & 0.186 & 0.358 & 0.458 & 0.409 & 0.312 \\
    Valley~\cite{luo2023valley} & 0.281 & 0.265 & 0.322 & 0.276 & 0.288 & 0.290 & 0.343 & 0.244 & 0.202 & 0.368 & 0.487 & 0.353 & 0.311 \\
    \bottomrule
    \end{tabular}
    }
    \end{threeparttable}
    \label{tab:exp_baseline1}
\end{table*}

\begin{table*}[!htb]
    \setlength{\abovecaptionskip}{1pt}
    \caption{BLEU score of baseline models in direct-answer setting.}
    \small
    \centering
    \begin{threeparttable}
    \resizebox{1\linewidth}{!}{
    \begin{tabular}{llllllllllllll}
    \toprule
    \multirow{2}{*}{Models} & \multicolumn{7}{c}{Questions concerning objects}        & \multicolumn{5}{c}{Questions concerning   actions} & \multirow{2}{*}{Overall} \\
    \cmidrule(lr){2-8}\cmidrule(lr){9-13}
                            & 1 & 2 & 3 & 4 & 5 & 6 & 7 & 8 & 9 & 10 & 11 & 12  &\\
    \cmidrule(lr){1-14}
    blind ChatGPT~\cite{OpenAI2023ChatGPT} & 0.032 & 0.053 & 0.062 & 0.038 & 0.062 & 0.009 & 0.010 & 0.035 & 0.133 & 0.046 & 0.105 & 0.036 & 0.052\\
    GPT4v~\cite{openai2023gpt4}          & 0.142 & 0.150 & 0.078 & 0.072 & 0.198 & 0.015 & 0.018 & 0.018 & 0.124 & 0.000 & 0.134 & 0.092 & 0.085 \\
    Video-LLaMa~\cite{zhang2023video} & 0.050 & 0.051 & 0.047 & 0.044 & 0.051 & 0.002 & 0.006 & 0.026 & 0.056 & 0.017 & 0.058 & 0.028 & 0.036 \\
    PandaGPT~\cite{su2023pandagpt}  & 0.089 & 0.096 & 0.085 & 0.082 & 0.097 & 0.008 & 0.014 & 0.054 & 0.103 & 0.035 & 0.123 & 0.052 & 0.070 \\
    Ask Anything~\cite{li2023videochat} & 0.141& 0.144& 0.110& 0.111& 0.174& 0.013& 0.030& 0.083& 0.127& 0.194& 0.202& 0.077& 0.110\\
    Video-ChatGPT~\cite{li2023videochat} & 0.142 & 0.125 & 0.108 & 0.090 & 0.152 & 0.014 & 0.044 & 0.079 & 0.143 & 0.109 & 0.164 & 0.067 & 0.096\\
    Valley~\cite{luo2023valley} & 0.099 & 0.096 & 0.095 & 0.084 & 0.107 & 0.010 & 0.012 & 0.070 & 0.112 & 0.035 & 0.134 & 0.048 & 0.068 \\
    \bottomrule
    \end{tabular}
    }
    \end{threeparttable}
    \label{tab:exp_baseline2}
\end{table*}

\begin{table*}[!htb]
    \setlength{\abovecaptionskip}{1pt}
    \caption{BERT score of baseline models in direct-answer setting.}
    \small
    \centering
    \begin{threeparttable}
    \resizebox{1\linewidth}{!}{
    \begin{tabular}{llllllllllllll}
    \toprule
    \multirow{2}{*}{Models} & \multicolumn{7}{c}{Questions concerning objects}        & \multicolumn{5}{c}{Questions concerning   actions} & \multirow{2}{*}{Overall} \\
    \cmidrule(lr){2-8}\cmidrule(lr){9-13}
                            & 1 & 2 & 3 & 4 & 5 & 6 & 7 & 8 & 9 & 10 & 11 & 12 &\\
    \cmidrule(lr){1-14}
    blind ChatGPT~\cite{OpenAI2023ChatGPT} & 0.887& 0.878& 0.892& 0.881& 0.884& 0.896& 0.887& 0.886& 0.883& 0.878& 0.892& 0.886& 0.886 \\
    GPT4v~\cite{openai2023gpt4}          & 0.954 & 0.953 & 0.955 & 0.959 & 0.956 & 0.955 & 0.953 & 0.963 & 0.954 & 0.957 & 0.952 & 0.957 & 0.956 \\
    Video-LLaMa~\cite{zhang2023video} & 0.961 & 0.960 & 0.961 & 0.961 & 0.961 & 0.961 & 0.961 & 0.961 & 0.961 & 0.961 & 0.961 & 0.961 & 0.961 \\
    PandaGPT~\cite{su2023pandagpt} & 0.952 & 0.953 & 0.952 & 0.952 & 0.953 & 0.952 & 0.953 & 0.952 & 0.952 & 0.952 & 0.952 & 0.952 & 0.952 \\
    Ask Anything~\cite{li2023videochat} & 0.957& 0.957& 0.958& 0.956& 0.961& 0.960& 0.955& 0.962& 0.959& 0.964& 0.966& 0.959& 0.959\\
    Video-ChatGPT~\cite{li2023videochat} & 0.955& 0.955& 0.957& 0.956& 0.957& 0.963& 0.952& 0.956& 0.958& 0.958& 0.961& 0.956& 0.957\\
    Valley~\cite{luo2023valley} & 0.955 & 0.955 & 0.955 & 0.955 & 0.954 & 0.954 & 0.955 & 0.955 & 0.955 & 0.955 & 0.954 & 0.955 & 0.955 \\
    \bottomrule
    \end{tabular}
    }
    \end{threeparttable}
    \label{tab:exp_baseline3}
\end{table*}

To evaluate the quality of our dataset, we ask two questions, which are answered in the following sections:
\begin{itemize}
    \item Do current methods effectively tackle our generated problems? Why do they perform better/worse compared with other QA datasets?
    \item Is our QA generation procedure efficient and justifiable?
\end{itemize}

\subsection{Baseline Results}

To address the first question, we select a range of recent video generation models, including Video-LLaMa~\cite{zhang2023video}, PandaGPT~\cite{su2023pandagpt}, Video-ChatGPT~\cite{li2023videochat}, AskAnything~\cite{li2023videochat}, and Valley~\cite{luo2023valley}. For deployable models, we specifically chose their Vicuna-7B-based versions~\cite{peng2023instruction}. Detailed baseline settings are described in the appendix.

\noindent\textbf{Baseline experiment settings.}
The baseline models are tested in two settings: ``multiple-choice'' and ``direct-answer''.
In the multiple-choice setting, models receive a question and four choice candidates; in the direct-answer setting, models are required to provide direct and succinct responses to the questions.
To calculate the accuracy in the multiple-choice setting, we need to parse an integer as the choice taken by the model. If the parsing algorithm fails, we calculate the BLEU score between model output and each choice candidate, picking the highest one as the model's answer. Results are shown in Table \ref{tab:exp_baseline1}.
We measure the BLEU score~\cite{papineni2002bleu} and BERT-F1 score~\cite{zhang2020bertscore} between the model's output and the generated answer in the ``direct-answer'' setting. Results are shown in Table \ref{tab:exp_baseline2} and \ref{tab:exp_baseline3} respectively.

\noindent\textbf{Result Analysis.} 
From the table, we have the following observations. First, all baseline models perform far from perfect.
GPT4v model has the best performance across different settings, which is consistent with its superior performance on standard leaderboards like~\cite{lu2023mathvista} and ~\cite{hendryckstest2021}. 
The best performing AskAnything (where ChatGPT is used as the language model) only has a BLEU score of 0.110 in table~\ref{tab:exp_baseline2}. It shows the value of the proposed benchmark to evaluate vision-language models' capabilities to understand and reason in videos.

Secondly, we notice a significant performance gap between API models (the GPT family) and deployable models (the LLaMa family with Vicuna-7B base model). We remark that this phenomenon can be attributed to the difference in both sizes and instruction-following abilities between GPT LLM family and LLaMa LLM family.

\begin{figure}[t]
  \setlength{\abovecaptionskip}{2pt}
  \setlength{\belowcaptionskip}{0pt}
  \centering
  \includegraphics[width=8cm]{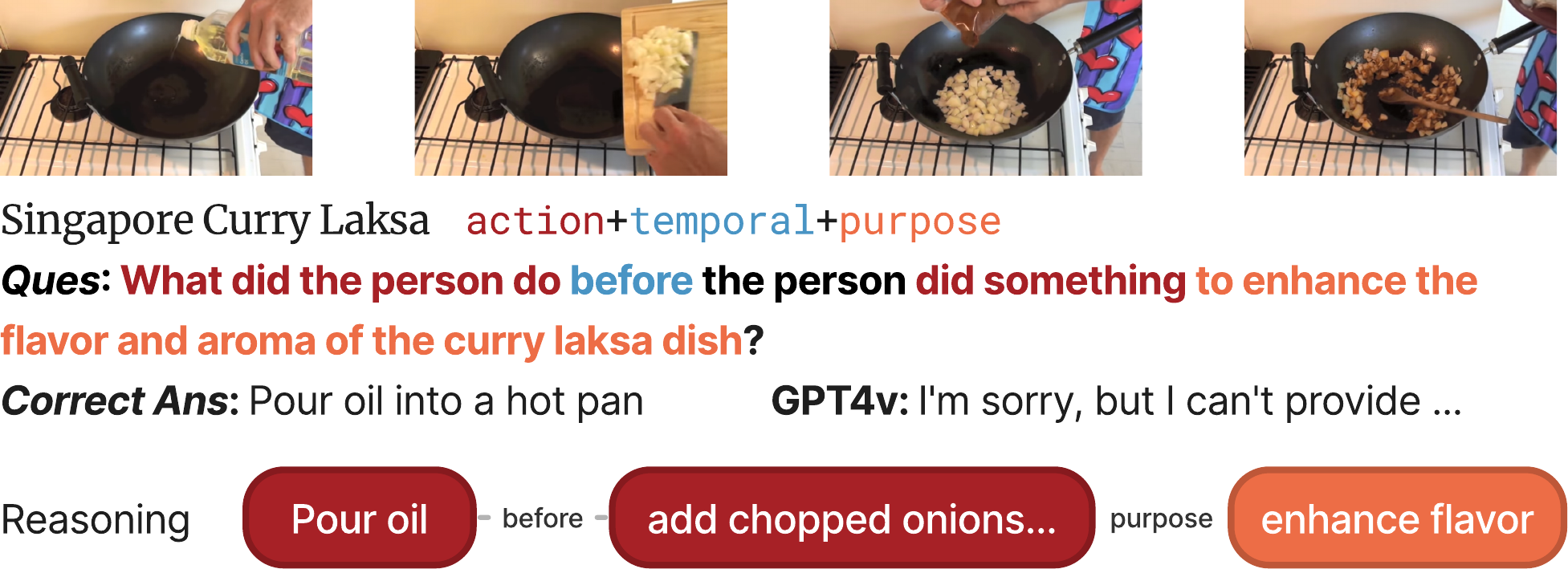}
  \caption{GPT4v's ability to perform complex combined spatiotemporal and situated commonsense reasoning is limited. The model needs to do two-hog reasoning, \ie, understanding the purpose of ``adding chopped onions'' is to ``enhance the flavor'' while knowing the previous action is ``pouring oil''.}
\label{fig:fail_case_of_gpt4v}
\end{figure}

Third, we find that most models perform better on simpler question types (\eg, OCT and APU) compared to more complex ones (\eg, OCB+ST and OCT+ST). For simpler questions, the choices might reveal the answers (\eg, Q: ``What did the person use to provide a sticky base to hold ingredients for California Roll?'', Choices: ``Rice'', ``Avocado'', ``Nori'', and ``Cucumber''). GPT4v may use this to achieve high multi-choice accuracy on OCB. However, GPT4v might struggle when both spatiotemporal and situated commonsense reasoning are required (see Figure~\ref{fig:fail_case_of_gpt4v}).

\label{sec:experiments_sc_vqa_dataset}





\subsection{Ablation Studies}

We answer the second question by conducting ablation studies to evaluate the effectiveness of each component within the generation method for the three graphs.

\noindent{\textbf{MLLMs for SKG Generation.}}
To justify the usage of MLLMs which provide spatiotemporal information, we focus on the four type QAs, namely ``Object Counterfactual'', ``Object Counterfactual + Spatiotemporal'', ``Object Contribution'', and ``Object Contribution + Spatiotemporal''. In the human evaluation, we ask whether the reasoning questions require to observe the video content, with results showing 91\%, 99\%, 94\%, and 99\% respectively. This confirms that MLLMs are critical to generating high-quality situated commonsense questions.


%

\noindent{\textbf{Effectiveness of Prompt Design in GKG Generation.}}
We use LLM to generate KGs in parsable JSON format. Here, we test the effectiveness of specifying output format and providing examples in the prompt structure for GKG generation. When given only the goal of the task, $0\%$ of the outputs are valid; specifying output format, this figure increases to $11.1\%$; with both output format and examples, all outputs can be parsed as JSON files.

\noindent{\textbf{Effectiveness of Few-Shot Self-Prompting.}}
We argue the few-shot examples of SCKG generation are necessary. By human testing, we evaluate the proportion of concrete and specific knowledge relations as opposed to un-situated, general knowledge. With Few-Shot Self-Prompting, $97\%$ of the generated knowledge is concrete while the ratio decreases to $61\%$ without the prompt technique.



\section{Conclusion}
\label{sec:conclusion}
Our research introduces a novel benchmark for Situated Open-World Commonsense Reasoning, advancing AI's ability to comprehend and reason in dynamic, real-world contexts. This benchmark includes a wide range of questions and situational analyses that go beyond traditional reasoning paradigms, challenging existing AI systems. Our novel approach in dataset generation offers scalability and enhanced logic in QA pair creation. While current models show promise, our experiment findings highlight the need for significant improvements, pointing towards exciting avenues for future general artificial intelligence.
\section{Acknowledgements}
\label{sec:acknowledgement}
We appreciate the contributions of our data quality validation team, Jiajing Zhang, Bingze Dai, Renxian Wang, Yue Xu, Yucong Li, Tuo Zhou, Wei Yi Oon, and Zijian Li.

\newpage
{
\bibliographystyle{ieeenat_fullname}
    \bibliography{main}
}

\end{document}